\documentclass[conference]{IEEEtran}

\usepackage{graphicx}
\usepackage{amsmath}
\usepackage{amssymb}
\usepackage{booktabs}
\usepackage{array}
\usepackage{cite}
\usepackage{url}

\title{Technical Report: One-Step Drifting Action Heads for GR00T N1.7}

\author{
    \IEEEauthorblockN{Xihe Shao}
    \IEEEauthorblockA{
        ZJU-UIUC Institute, Zhejiang University\\
        Hangzhou, China\\
        Intern, LimX Dynamics, Shenzhen, China\\
        xihe.25@intl.zju.edu.cn
    }
}

\date{September 2026}

\begin{document}

\maketitle

\begin{abstract}
One-step action generation can substantially reduce the inference cost of
vision-language-action (VLA) policies, but its effect on closed-loop task
success remains an open question. This technical report studies a GR00T N1.7
variant in which the iterative diffusion-transformer action head is replaced
by a one-step drifting action head, together with an overlap-conditioned
extension for asynchronous chunk replacement. All multi-seed drifting runs
were trained on two NVIDIA A800 GPUs. On LIBERO, the action head reduces the
mean model-forward time of the action head from approximately
$45.3\,\mathrm{ms}$ to $5.0\,\mathrm{ms}$, while the measured
backbone-plus-head time falls from approximately $70.0\,\mathrm{ms}$ to
$30.6\,\mathrm{ms}$. However, this speedup is accompanied by a systematic
reduction in task success. Across three drifting seeds, success is
$64.0\pm4.0\%$ on LIBERO-Spatial, $52.0\pm1.0\%$ on LIBERO-Goal, and
$26.0\pm2.6\%$ on LIBERO-Long. The low seed variance indicates that the
degradation is not explained by random initialization alone. We report the
result as a speed--success trade-off rather than an overall improvement, and
discuss likely contributing factors including deterministic one-step mode
averaging, batch-dependent geometry estimation, long open-loop chunk
execution, and the fact that synchronous LIBERO evaluation does not exercise
the asynchronous overlap path.
\end{abstract}

\begin{IEEEkeywords}
vision-language-action models, action chunking, one-step generation,
drifting models, inference acceleration, robot manipulation
\end{IEEEkeywords}

\section{Introduction}

Vision-language-action (VLA) models have demonstrated strong capabilities in
robotic manipulation, but their computational cost and inference latency can
limit deployment on real-time robot platforms. Existing efficiency methods
reduce different parts of the VLA pipeline. Model-level approaches include
fast--slow architectures~\cite{chen2026fast}, lightweight backbones
\cite{wen2025tinyvla}, token caching~\cite{xu2026vla}, and adaptive
early-exit inference~\cite{yue2024deer}. System-level approaches include
quantization~\cite{park2024quantization}, graph optimization, and custom
operators~\cite{FluxVLA2026}. Iterative action generation can nevertheless
remain an important source of latency because diffusion and flow-matching
heads require multiple network evaluations per action chunk.

This report studies whether the iterative action head of GR00T
N1.7~\cite{gr00tn1_2025} can be replaced by a one-step drifting action head.
The base training objective is inherited from Implicit Drifting Policy
(IDP)~\cite{yang2026implicitdrifting}; this report does not claim a new
generative objective. The implemented head retains the GR00T
vision-language backbone and preprocessing stack, replaces the iterative
flow-matching action generator with a deterministic conditional transformer,
and predicts a complete action chunk in one action-head evaluation.

A second issue appears in asynchronous action prefetching. If a replacement
chunk is generated while the current chunk is still being executed, the new
prediction may not account for commands that have already been committed to
the controller. To address this issue, we implement an overlap-conditioned
variant, called DrifOv, inspired by training-time real-time
chunking~\cite{black2025training}. DrifOv conditions on a valid action prefix,
an overlap length, and an estimated execution offset, and copies the committed
prefix exactly while generating only the unknown suffix.

The report has both positive and negative findings. The positive result is a
large and highly stable action-head speedup. The negative result is that this
speedup does not currently preserve task success: under the evaluated training
protocol, the drifting variants are substantially less successful than the
GR00T reference on three LIBERO suites. The three drifting seeds show low
variance, so the gap is systematic rather than attributable only to a single
unlucky initialization.

The contributions of this technical report are:
\begin{itemize}
    \item We implement and describe a one-step drifting action head for
    GR00T N1.7, including its inherited training objective, model dimensions,
    and one-evaluation inference path.
    \item We implement an overlap-conditioned extension that represents an
    already committed action prefix, its coordinate-wise validity, the overlap
    length, and an estimated execution offset while preserving valid prefix
    coordinates exactly.
    \item We provide a multi-seed LIBERO study in which all drifting seed runs
    use the same two-A800 training setup. We report latency and success
    separately and do not average suites with missing measurements.
    \item We identify several hypotheses for the observed speed--success
    trade-off and specify the experiments needed to isolate them. The current
    evidence does not establish that the drifting objective itself causes the
    entire degradation.
    \item We record preliminary real-robot deployment and data-collection
    experience. These experiments are presented as engineering observations,
    not as controlled policy comparisons.
\end{itemize}

This is a preliminary technical report, not a claim of state-of-the-art
performance. The timing measurements include only the VLM backbone and the
action head. They exclude preprocessing, postprocessing, networking, action
transmission, controller execution, and scheduling. The reported speedups
therefore should not be interpreted as equal reductions in end-to-end robot
latency.

\section{Related Work}

\subsection{Efficient VLA Inference}
VLA efficiency has been approached by reducing the size or frequency of VLM
computation. TinyVLA~\cite{wen2025tinyvla} uses a compact vision-language
backbone for data-efficient control. VLA-Cache~\cite{xu2026vla} reuses
vision-language tokens across an action trajectory. Fast-in-Slow
\cite{chen2026fast} separates slow reasoning from fast action generation, and
DEER~\cite{yue2024deer} dynamically terminates reasoning when further
computation is not needed. Quantization and optimized execution engines
\cite{park2024quantization,FluxVLA2026} reduce the cost of the remaining
network evaluations.

These methods are complementary to the action-head replacement studied here.
In the GR00T N1.7 configuration considered in this report, the VLM backbone
dominates the measured model-forward time even after the action head is
reduced to approximately $5\,\mathrm{ms}$. For this reason, an action-head-only
latency reduction cannot by itself make the complete pipeline real-time.

\subsection{Action Chunking and Real-Time Chunking}
Action-chunking policies predict multiple future actions in one policy call
and execute a queue of actions before the next request. This reduces VLM calls
but creates a trade-off between open-loop execution length and feedback
frequency. Real-time chunking (RTC)~\cite{black2025training} begins the next
request before the current chunk is exhausted and constrains a new chunk to
remain consistent with the unexecuted actions that will execute while
inference is in flight.

DrifOv follows the same asynchronous timeline but does not use iterative
inference-time guidance. Instead, it learns prefix-conditioned one-step
generation. Committed coordinates are copied exactly after decoding, while the
suffix is generated jointly in a single action-transformer evaluation.

\subsection{Drifting Models and One-Step Generation}
The base action head uses the training objective of Implicit Drifting Policy
(IDP)~\cite{yang2026implicitdrifting}. For each observation-action pair, IDP
constructs a geometry-weighted potential around the expert action instead of
regressing an explicit continuous drifting field. At deployment, a zero action
seed is mapped directly to a complete action chunk.

The extension in this report is not the drifting objective itself. The
implementation adds an action-prefix interface and overlap-aware suffix
generation so that a successor chunk can be conditioned on commands already
committed to the controller. The report evaluates the implementation as an
engineering replacement for the GR00T flow-matching action head.

\section{Method}

\subsection{Scope and Architecture}
The implementation retains the GR00T N1.7 preprocessing and
Cosmos-Reason2/Qwen3 vision-language backbone. Hidden features are read from
layer 16 and projected to the action-head dimension. Robot state is encoded by
an embodiment-specific MLP, and actions are encoded by an embodiment-specific
action encoder. The action head is a 12-block transformer with model dimension
$1024$, $16$ attention heads, feed-forward dimension $4096$, and dropout
$0.1$. Each block contains self-attention over action and state tokens,
cross-attention to vision-language memory, and a feed-forward network. The
cross-attention alternates between image and text tokens across blocks.

The default action chunk length is $H=40$. The maximum state and action
dimensions are both $132$, and the state history length is one. At deployment,
the head receives a zero action seed and produces the complete chunk in one
action-transformer evaluation. The number of inference steps is fixed to one.

\subsection{Inherited Drifting Objective}
\label{sec:inherited_objective}

The base action head uses the IDP objective~\cite{yang2026implicitdrifting}.
For an observation-action pair $(o_i,a_i^*)$, let $M_i\succeq0$ be a
detached diagonal geometry matrix. The potential is
\begin{equation}
E_i(a)
=
\frac{1}{2}
(a-a_i^*)^\top (I+M_i)(a-a_i^*).
\label{eq:potential}
\end{equation}
The corresponding correction is
\begin{equation}
-\nabla_a E_i(a)=(I+M_i)(a_i^*-a).
\label{eq:correction}
\end{equation}

Training evaluates the network at two locations. The proposal prediction
starts from a zero action seed at time zero:
\begin{equation}
y_i=f_\theta(o_i,\mathbf{0},0).
\label{eq:proposal}
\end{equation}
The proximal prediction starts near the expert action:
\begin{equation}
\tilde a_i=a_i^*+(1-t_*)\epsilon_i,
\qquad
\epsilon_i\sim\mathcal{N}(0,I),
\label{eq:proximal_seed}
\end{equation}
\begin{equation}
z_i=f_\theta(o_i,\tilde a_i,t_*),
\label{eq:proximal_prediction}
\end{equation}
with $t_*=0.9$ in the reported configuration. The inherited objective is
\begin{equation}
\mathcal{L}_{\mathrm{IDP}}
=
E_i(y_i)+\lambda_{\mathrm{prox}}E_i(z_i),
\label{eq:idp_loss}
\end{equation}
where $\lambda_{\mathrm{prox}}=81.0$ in the reported implementation.

\subsection{Geometry Weighting}
For a minibatch of observation-action pairs, normalized observation features
are compared by inner product. The similarities are row-standardized and
converted to neighbor weights $w_{ij}$ with a softmax. For action coordinate
$d$, the conditional variance is estimated as
\begin{equation}
v_{i,d}^{\mathrm{cond}}
=
\sum_j w_{ij}(a_{j,d}^*-a_{i,d}^*)^2.
\label{eq:conditional_variance}
\end{equation}
The reference variance $v_d^{\mathrm{ref}}$ is computed from the same
minibatch. Normalized inverse variances are compared to form the diagonal
geometry excess
\begin{equation}
m_{i,d}
=
\operatorname{ReLU}
\left(
\frac{s_{i,d}^{\mathrm{cond}}}{s_d^{\mathrm{ref}}+\varepsilon}
-1
\right),
\qquad
M_i=\operatorname{Diag}(m_i),
\label{eq:geometry_excess}
\end{equation}
where $s^{\mathrm{cond}}$ and $s^{\mathrm{ref}}$ are normalized precision
scales. The geometry matrix is detached and used only to weight the training
potential.

Two implementation details are relevant to the interpretation of the
experiments. First, the reference variance is estimated within the current
minibatch, so its quality depends on batch size and batch composition. Second,
the current implementation includes the query sample in the neighbor set,
although the IDP equation excludes $j=i$. This report treats these choices as
limitations rather than validated design decisions.

\subsection{Overlap-Conditioned Suffix Generation}
\label{sec:prefix_conditioning}

DrifOv extends the one-step head with an explicit action prefix. Let
\begin{equation}
m_{h,j}
=
\mathbb{I}[h<L]\nu_{h,j},
\label{eq:mask}
\end{equation}
where $h$ indexes action-chunk positions, $j$ indexes action coordinates,
$L$ is the number of committed prefix rows, and $\nu_{h,j}$ is the valid-action
mask. The prefix rows $p_{0:L-1}$ are supplied to the head together with a
prefix-validity mask, the overlap length $L$, and an execution offset.

The action query for output position $h$ is formed from:
\begin{enumerate}
    \item the encoded masked prefix,
    \item a learned prefix-mask embedding,
    \item an overlap-length embedding,
    \item an execution-offset embedding, and
    \item a positional embedding.
\end{enumerate}
The decoder jointly processes all $H$ action queries and cross-attends to the
vision-language and state-conditioning tokens. For each valid prefix
coordinate, the decoded value is replaced by the supplied prefix value:
\begin{equation}
\widehat A_{h,j}
=
m_{h,j}p_{h,j}
+
(1-m_{h,j})\widetilde A_{\theta,h,j}.
\label{eq:hard_prefix_copy}
\end{equation}
Thus, committed actions cannot be rewritten by the learned head. The prefix
still influences the suffix through self-attention. The model is trained to
generate only the unknown suffix.

\subsection{Overlap-Aware Training}
\label{sec:overlap_training}

For each training action chunk, an overlap length is sampled from zero through
$H-1$. The reported configuration also samples a zero-overlap example with
probability $0.2$. A non-empty prefix is copied from the expert action chunk.
The implementation then perturbs the prefix with probability $0.5$ using
Gaussian noise with standard deviation $0.02$ and clips the perturbation to
$[-0.05,0.05]$. The execution offset is sampled between zero and the selected
overlap length. Prefix conditioning can be disabled for an unconditioned
baseline.

The generation loss is the inherited potential evaluated only on suffix
coordinates. Let
\begin{equation}
\omega_{h,j}
=
\nu_{h,j}(1-m_{h,j})
\label{eq:suffix_mask}
\end{equation}
select the coordinates for which the model is responsible. The suffix loss is
\begin{equation}
\mathcal{L}_{\mathrm{gen}}
=
\frac{
\sum_{h,j}\omega_{h,j}E_i(\widehat A_{h,j})
}{
\sum_{h,j}\omega_{h,j}
}.
\label{eq:suffix_loss}
\end{equation}
The same objective is applied to the proximal prediction, with prefix
coordinates replaced by the supplied prefix before loss evaluation.

The implementation optionally adds boundary terms at the first generated
suffix action. In the reported configuration, the weights are $1.0$ for the
boundary action error and $0.1$ for the finite-difference velocity error. The
acceleration weight is zero. These terms are intended to encourage continuity
between the fixed prefix and the generated suffix. They do not turn the
problem into a new generative model; they are auxiliary regularizers.

\subsection{One-Step Inference}
\label{sec:inference}

For synchronous LIBERO evaluation, no action prefix is supplied. The inference
path therefore uses an empty prefix, mask, and overlap length. The action seed
is zero, the time input is zero, and the model produces one complete action
chunk in one action-head evaluation:
\begin{equation}
\widehat A
=
f_\theta(o,s,\mathbf{0},0).
\label{eq:one_step_inference}
\end{equation}
This is the path measured in the LIBERO experiments in
Section~\ref{sec:libero_results}. It tests the unconditioned one-step head and
does not test asynchronous overlap conditioning.

For asynchronous deployment, the runtime can supply the remaining normalized
action rows, their true length before padding, and an estimated inference
delay. The committed rows are returned unchanged and only the unknown suffix is
generated. A request is rejected if the estimated delay exceeds the available
prefix rather than executing a suffix conditioned on an expired prefix.

\subsection{Relationship to the Proposed History-Latent Formulation}
An earlier design note proposed a causal history-memory branch and a
delay-propagator that would update the latent state using actions executed
while inference is in flight. That formulation is not implemented or
evaluated in this report. The implemented method conditions only on the
committed action prefix and global overlap metadata. The distinction is
important: the experiments below do not establish the benefit of a
history-latent model.

\section{Experiments}
\label{sec:experiments}

\subsection{Experimental Protocol}

The LIBERO experiments use the LIBERO benchmark \cite{liu2023libero},
the \texttt{lerobot/libero} dataset, and the LeRobot training and evaluation
stack. All multi-seed drifting checkpoints are trained
on two NVIDIA A800 GPUs. The documented seed-1000 run uses batch size 32 per
GPU, 20k optimization steps, and bfloat16 precision. The same named recipe is
used for the archived seed-1, seed-42, and seed-1000 drifting runs. The
primary comparison against GR00T uses the seed-1 and seed-42 runs from the
archived seed-comparison directory. The seed-1000 GR00T checkpoint comes from
an older archived evaluation and is reported only as a reference point.

The success summaries use 10 episodes per LIBERO subtask, corresponding to
100 rollouts per suite. Latency is recorded by timing hooks around the
backbone and the action head. Each hook performs a CUDA synchronization before
and after the timed region. The reported values are means over all inference
calls within a suite. They exclude preprocessing, postprocessing, networking,
action transmission, and low-level control, and are therefore not end-to-end
deployment latencies.

\begin{table}[t]
\centering
\small
\caption{Configuration of the reported experiments.}
\label{tab:setup}
\begin{tabular}{p{0.34\linewidth}p{0.56\linewidth}}
\toprule
Item & Configuration \\
\midrule
Training hardware & $2\times$ NVIDIA A800 \\
Deployment hardware & RTX 4090 + Intel 14900KF \\
Dataset & \texttt{lerobot/libero} \\
Training steps & 20k \\
Batch size & 32 per GPU, effective 64 \\
Precision & bfloat16 \\
Action chunk & $H=40$ \\
LIBERO episodes & 10 per subtask, 100 per suite \\
Drifting seeds & 1, 42, 1000 \\
\bottomrule
\end{tabular}
\end{table}

\subsection{LIBERO Latency and Success}
\label{sec:libero_results}

Table~\ref{tab:latency} reports the measured latency. The action head is
approximately $9\times$ faster across all suites, while the full measured
model-forward path is approximately $2.2$--$2.4\times$ faster. The VLM
backbone remains nearly unchanged, confirming that action-head replacement
addresses only one component of the inference pipeline.

\begin{table*}[t]
\centering
\small
\caption{Measured model-forward latency on LIBERO. Values are means over the
recorded seeds and inference calls. ``Total'' is backbone plus action head.
The measurements exclude preprocessing, postprocessing, and control overhead.}
\label{tab:latency}
\begin{tabular}{lcccccc}
\toprule
Task & Model & Seeds & Backbone (ms) & Action head (ms) & Total (ms) & Head speedup \\
\midrule
Spatial & GR00T N1.7 & 3 & $24.99\pm0.28$ & $45.36\pm0.17$ & $70.35\pm0.19$ & $1.0\times$ \\
Spatial & DrifOv & 3 & $24.42\pm0.11$ & $4.96\pm0.08$ & $29.38\pm0.05$ & $9.1\times$ \\
Goal & GR00T N1.7 & 3 & $25.03\pm0.45$ & $45.43\pm0.19$ & $70.46\pm0.50$ & $1.0\times$ \\
Goal & DrifOv & 3 & $26.47\pm3.69$ & $5.07\pm0.30$ & $31.54\pm3.97$ & $9.0\times$ \\
Long & GR00T N1.7 & 2 & $24.14\pm0.43$ & $45.18\pm0.07$ & $69.33\pm0.36$ & $1.0\times$ \\
Long & DrifOv & 3 & $25.92\pm3.46$ & $5.04\pm0.27$ & $30.95\pm3.72$ & $9.0\times$ \\
Object & GR00T N1.7 & 3 & $26.21\pm1.42$ & $45.80\pm0.12$ & $72.01\pm1.51$ & $1.0\times$ \\
Object & DrifOv & 3 & $24.30\pm0.14$ & $4.94\pm0.04$ & $29.25\pm0.16$ & $9.3\times$ \\
\bottomrule
\end{tabular}
\end{table*}

Table~\ref{tab:success} reports success. The new drifting seed runs show a
large and systematic gap relative to the GR00T reference. The low standard
deviation across the three drifting seeds is important: the gap is not
explained by one failed initialization.

\begin{table*}[t]
\centering
\small
\caption{LIBERO success rate. The multi-seed drifting results use seeds 1,
42, and 1000. GR00T uses seeds 1 and 42 for the matched seed-comparison runs;
the GR00T spatial, goal, and Long values also include an older seed-1000
reference checkpoint. Object success is unavailable for the drifting seed
runs and is not averaged.}
\label{tab:success}
\begin{tabular}{lccccc}
\toprule
Task & Model & Seeds & Success (\%) & Per-seed success (\%) & Coverage \\
\midrule
Spatial & GR00T N1.7 & 3 & $82.3\pm6.8$ & 77, 90, 80 & 3/3 \\
Spatial & DrifOv & 3 & $64.0\pm4.0$ & 60, 64, 68 & 3/3 \\
Goal & GR00T N1.7 & 3 & $82.3\pm4.0$ & 80, 87, 80 & 3/3 \\
Goal & DrifOv & 3 & $52.0\pm1.0$ & 52, 51, 53 & 3/3 \\
Long & GR00T N1.7 & 2 & $74.5\pm10.6$ & 67, 82 & 2/2 \\
Long & DrifOv & 3 & $26.0\pm2.6$ & 28, 27, 23 & 3/3 \\
Object & GR00T N1.7 & 1 & 88.0 & 88 & 1/3 \\
Object & DrifOv & 0 & N/A & N/A & 0/3 \\
\bottomrule
\end{tabular}
\end{table*}

\begin{figure*}[t]
\centering
\includegraphics[width=0.92\textwidth]{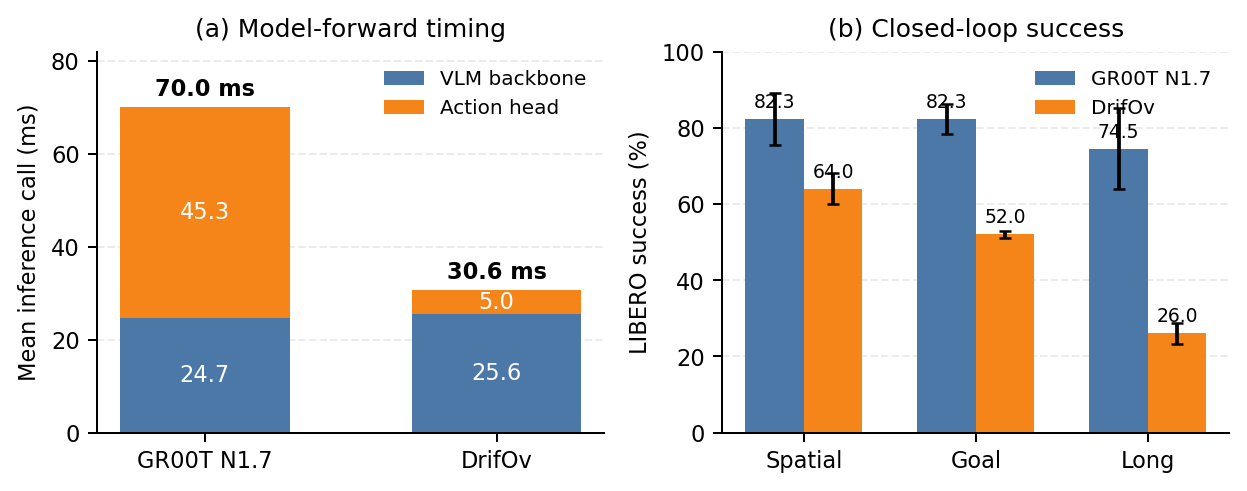}
\caption{Summary of the speed--success trade-off. Panel (a) averages the
measured model-forward time over Spatial, Goal, and Long. Panel (b) reports
success for the three suites with complete or partially matched summaries.
Error bars in panel (b) are standard deviations across the available training
seeds.}
\label{fig:main_summary}
\end{figure*}

The three drifting seeds produce similar success within each suite:
Spatial $60/64/68\%$, Goal $52/51/53\%$, and Long $28/27/23\%$. In contrast
to the near-constant latency, the loss of task success grows with task
horizon. The largest degradation appears on LIBERO-Long, where the policy must
sustain progress over a longer sequence of action chunks.

\subsection{Legacy LIBERO Profile}
\begin{figure}[t]
\centering
\includegraphics[width=\linewidth]{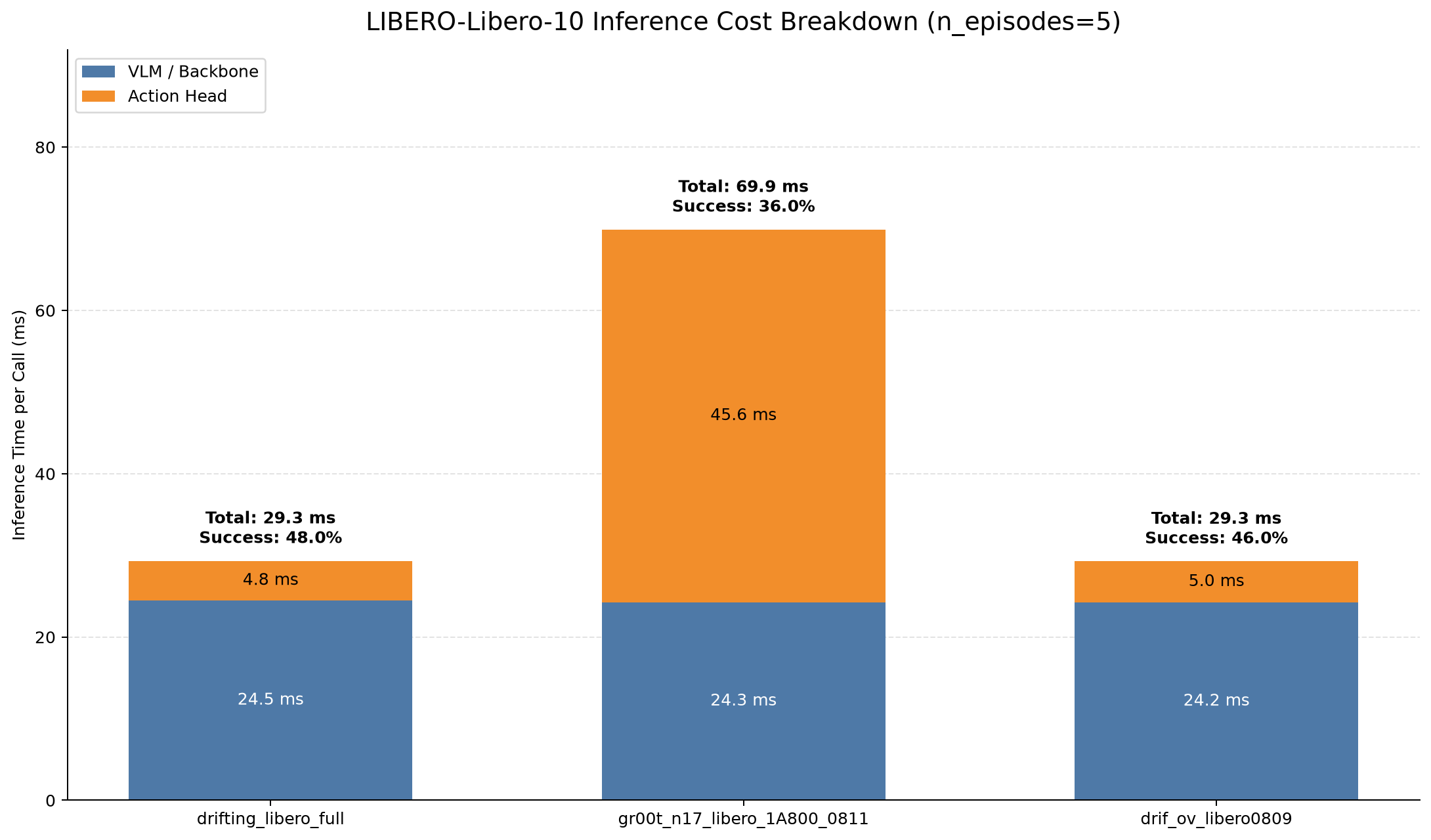}
\caption{Legacy LIBERO-Long profile from an earlier checkpoint. The legacy
run is shown as an engineering record only; it is not the primary two-A800
seed comparison and is not used to estimate the method's success.}
\label{fig:legacy}
\end{figure}

An earlier single-run study used a different checkpoint and training regime.
Those measurements are not part of the primary comparison. Figure~\ref{fig:legacy}
reproduces the legacy profile only to document the development history. Its
success values should not be compared directly to the multi-seed results.

\subsection{Training Memory Observation}
\begin{figure*}[t]
\centering
\begin{minipage}{\textwidth}
\centering
\includegraphics[width=0.315\textwidth]{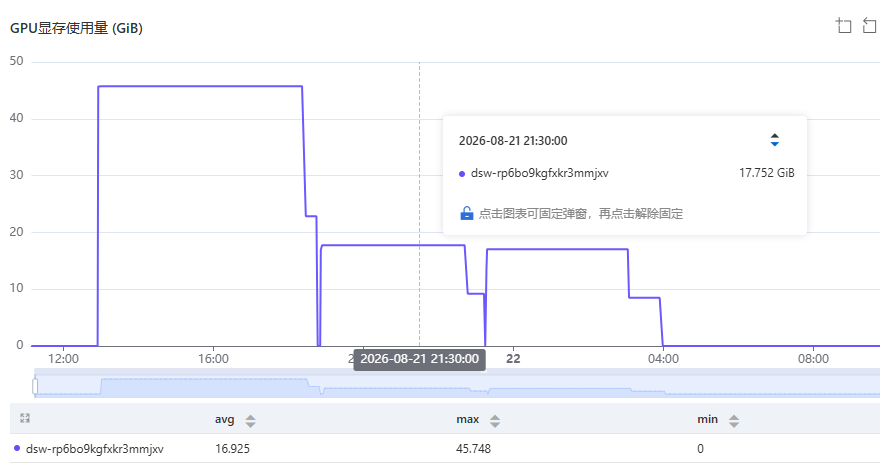}\hfill
\includegraphics[width=0.315\textwidth]{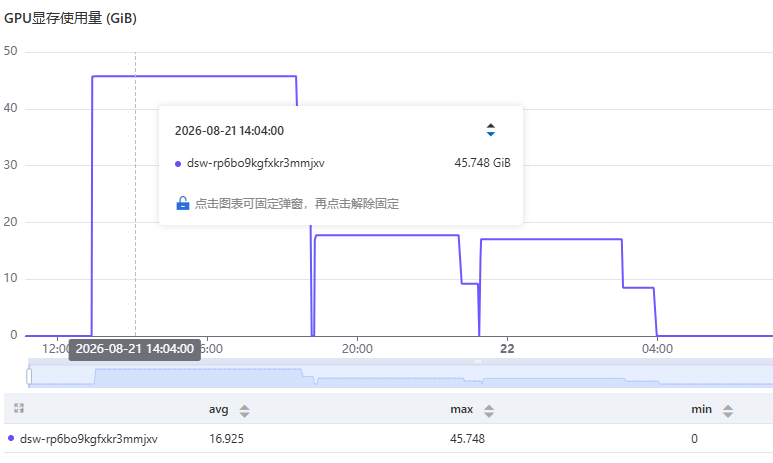}\hfill
\includegraphics[width=0.315\textwidth]{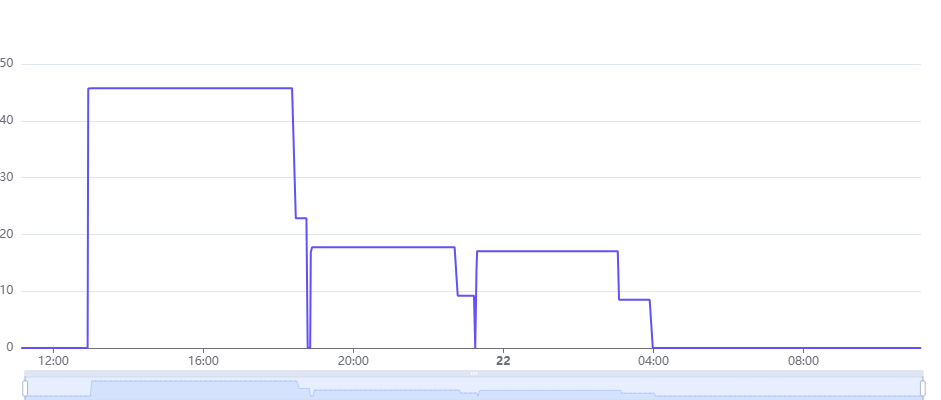}

\vspace{2pt}
\makebox[0.315\textwidth]{\small (a) Drifting}
\hfill
\makebox[0.315\textwidth]{\small (b) GR00T N1.7}
\hfill
\makebox[0.315\textwidth]{\small (c) Combined telemetry}
\end{minipage}
\caption{GPU-memory telemetry recorded during the drifting and GR00T training
runs on the same server. The drifting run shows a lower selected peak than the
GR00T run. These measurements are anecdotal rather than a controlled benchmark,
because the telemetry also includes other jobs and idle periods.}
\label{fig:memory}
\end{figure*}

Figure~\ref{fig:memory} shows GPU-memory telemetry recorded while running the
GR00T and drifting training jobs on the same server. Selected peaks are
$45.748\,\mathrm{GiB}$ for the GR00T run and $17.752\,\mathrm{GiB}$ for the
drifting run. This observation is anecdotal rather than a controlled memory
benchmark. We therefore do not attribute the difference solely to the action 
head, but record it as a potential practical benefit for resource-limited training.

\subsection{Preliminary Real-Robot Deployment}
The policy stack was deployed on a LimX Tron2 robot with an RTX 4090 and Intel
14900KF workstation. We collected and annotated data for three manipulation
settings. These experiments demonstrate system integration and data
collection, but they are not controlled success-rate comparisons.

\begin{figure}[t]
\centering
\includegraphics[width=\linewidth]{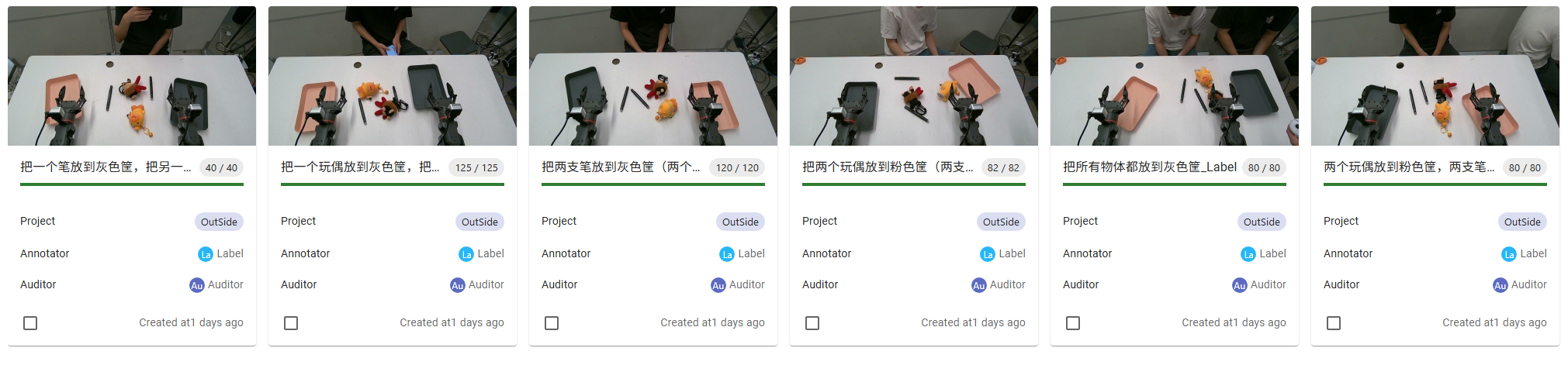}
\caption{Real-robot toy-placement and bin-sorting task variants used during
data collection.}
\label{fig:place}
\end{figure}

\paragraph{Task 1: toy placement.}
The first setting places toys and writing tools into specified bins. The
annotation interface used for the task variants is shown in
Fig.~\ref{fig:place}. Success and failure labels are recorded for data
curation, but the collected labels were not collected under a fixed policy
evaluation protocol.

\begin{figure}[t]
\centering
\includegraphics[width=0.72\linewidth]{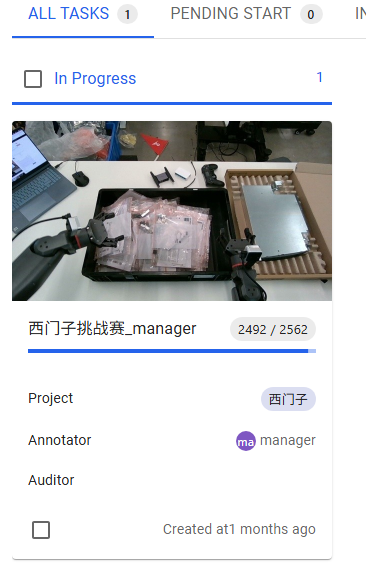}
\caption{Annotation progress for the industrial accessory-packaging task. A
total of 2,562 rollout trajectories were collected.}
\label{fig:package}
\end{figure}

\paragraph{Task 2: industrial accessory packaging.}
The second setting places packaged industrial accessories into a target tray.
A total of 2,562 annotated rollout trajectories were collected. A representative
interface is shown in Fig.~\ref{fig:package}. The dataset is substantially
larger than the exploratory LIBERO seed runs, but no matched policy comparison
was completed on this task.

\begin{figure}[t]
\centering
\includegraphics[width=\linewidth]{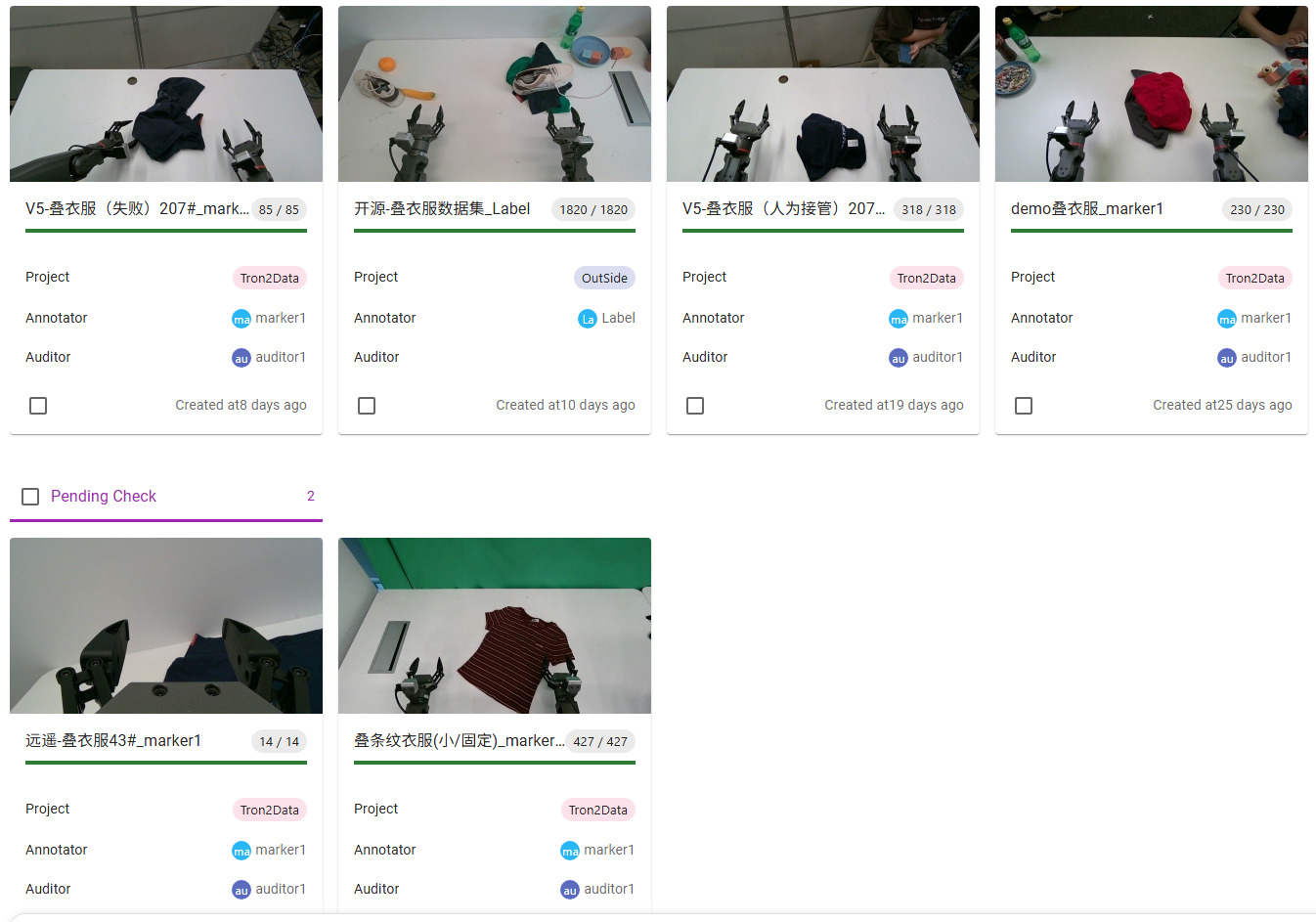}
\caption{Cloth-folding datasets collected during real-robot deployment,
including successful, failed, and human-takeover demonstrations.}
\label{fig:cloth}
\end{figure}

\paragraph{Task 3: cloth folding.}
The third setting contains cloth-folding data, including successful
demonstrations, failed attempts, and human takeover segments. The annotation
overview is shown in Fig.~\ref{fig:cloth}. These data were collected to study
deformable-object manipulation, but the reported technical snapshot does not
contain a controlled policy-success comparison.

\subsection{Results Excluded from the Primary Comparison}
The pi0.5 evaluation reported $0.0\%$ success on every LIBERO suite in the
archived timing directory. Because an all-zero result across suites is more
consistent with an evaluation or action-mapping failure than a policy result,
it is excluded from the primary comparison. The timing CSV and summary JSON
also disagree for several pi0.5 tasks.

The legacy \texttt{drif\_ov\_libero0809} checkpoint is excluded from the
controlled comparison for two reasons. First, it was trained with four A800
GPUs, batch size 256 per GPU, and 15k steps, whereas the multi-seed checkpoints
use the two-A800 20k-step recipe. Second, its reported success comes from a
single exploratory run. The legacy results are retained in the appendix as a
development record, not as evidence of method quality.

\section{Discussion}
\label{sec:discussion}

\subsection{What the Current Results Establish}
The results establish a large and repeatable model-forward speedup. The action
head is approximately nine times faster, and the backbone-plus-head path is
approximately $2.2$--$2.4$ times faster. The speedup is stable across seeds
and suites. The VLM backbone remains the dominant component after the
replacement, so the measured improvement does not make the complete
vision-language-action pipeline real-time by itself.

The results also establish a negative outcome. Under the evaluated two-A800
training recipe, the drifting seeds are consistently less successful than the
GR00T reference on Spatial, Goal, and Long. The failure is largest on
LIBERO-Long. Because the seed variance is small, random initialization is not
a sufficient explanation.

The experiments do not isolate which change causes the degradation. The
following hypotheses are therefore stated as alternatives rather than
conclusions.

\subsection{Deterministic One-Step Mode Averaging}
The inherited objective applies an isotropic error plus a coordinate-wise
geometry penalty. This is still a regression target toward a demonstration
action. In a one-step deterministic generator, observations with multiple
valid modes can produce a prediction near the conditional mean of those
modes. Near the mean, no demonstrated trajectory may exist. A multi-step
diffusion or flow generator can move between modes over its trajectory, while
the one-step model must commit to one location immediately.

The action-chunk space is high-dimensional and may contain several valid
continuations even when each individual action is roughly unimodal. This
mechanism is consistent with the increase in failure as the task horizon
grows, but the current experiments do not measure mode coverage directly.
Useful diagnostics would compare one-step prediction error as a function of
the number of expert modes near each observation and evaluate whether a latent
or discrete mode variable reduces the error.

\subsection{Batch-Dependent Geometry Estimation}
The geometry matrix is estimated from the current minibatch. The reported
two-A800 runs use 32 samples per local batch. Neighbor weights, conditional
variances, and reference variances therefore vary with batch composition.
Changing batch size changes the effective geometry objective even when the
optimization objective and data are otherwise matched.

The current implementation also includes the query sample in the neighbor
set, while the IDP equation excludes the self-neighbor term. Inclusion of the
diagonal similarity can make the softmax neighborhood more concentrated and
can alter the variance estimate. This report does not claim that either choice
is correct; both require ablation. A retrieval-based or dataset-level geometry
estimate would remove the direct dependence on minibatch size.

\subsection{Open-Loop Chunk Execution}
The policy predicts $H=40$ actions, and the default evaluation executes the
chunk before requesting the next one. A faster action head does not
automatically reduce this open-loop interval. Long-horizon error can therefore
accumulate even though the head itself is fast.

A direct engineering experiment is to vary the number of actions executed
before replanning. Candidate settings are $1$, $2$, $5$, $10$, $20$, and $40$
actions. If task success improves as the execution horizon decreases, the
reported latency advantage can be translated into higher feedback frequency
rather than being spent on a fixed long chunk. This experiment is not included
in the current snapshot.

\subsection{Untested Asynchronous Overlap Path}
The DrifOv training and inference code supports a committed prefix, but the
synchronous LIBERO evaluation supplies an empty prefix. The measured LIBERO
success therefore evaluates the unconditioned one-step path. It does not
measure whether prefix conditioning improves chunk handoff, queue continuity,
or asynchronous control.

The training prefix is sampled from the expert chunk and optionally perturbed
with bounded Gaussian noise. During asynchronous deployment, however, the
prefix would be produced by the policy itself. The training distribution and
the deployment distribution are not identical. Model-generated prefixes with
physically consistent arrival states would be required to evaluate this
mechanism properly.

\subsection{Training and Architecture Confounds}
Both GR00T and DrifOv retain the same backbone and preprocessing stack, but
the action-head architecture, action-head initialization, and training
objective differ. The current study does not include a direct MSE head, an
ablation of the geometry term, or a comparison against a one-step
consistency-distilled GR00T head. The low LIBERO scores may therefore reflect
optimization difficulty, an architecture mismatch, an action-normalization
issue, or the one-step objective rather than a single isolated factor.

The report also does not use a pretrained GR00T action head as the
initialization for the new head. The reported comparison is consequently a
from-scratch action-head comparison rather than a fine-tuning comparison.

\subsection{Implications for Deployment}
The observed memory telemetry and the large action-head speedup suggest that
one-step heads are worth continuing to investigate for resource-constrained
systems. The current LIBERO evidence, however, does not support a claim that
the action head can replace the iterative head without performance loss.

A realistic next system design is a slow VLM backbone combined with a faster
action head. The backbone could be refreshed at a lower rate while the action
head is recomputed from cached visual features and the latest robot state.
That design requires delay augmentation and arrival-state consistency
training; it is not evaluated here.

\subsection{Threats to Validity}
The following factors limit the conclusions of this report:
\begin{itemize}
    \item The GR00T reference checkpoint for seed 1000 comes from an older
    archived evaluation and may not use the exact same protocol as the
    seed-comparison runs.
    \item Object-suite success is missing for the drifting seeds. The object
    timing row is not accompanied by a controlled success comparison.
    \item The pi0.5 results are excluded because the archived evaluation
    produced zero success on every suite and inconsistent summary files.
    \item Timing includes only the backbone and action head. It excludes
    preprocessing, networking, controller execution, and scheduling.
    \item The real-robot section records data collection and system
    integration, but does not provide a controlled success-rate comparison.
    \item The memory observation comes from shared server telemetry and is
    affected by other jobs.
\end{itemize}

\section{Conclusion}

This technical report studied a one-step drifting replacement for the
iterative action head of GR00T N1.7. The implementation reduces the measured
action-head time from approximately $45.3\,\mathrm{ms}$ to $5.0\,\mathrm{ms}$
and the measured backbone-plus-head time from approximately $70.0\,\mathrm{ms}$
to $30.6\,\mathrm{ms}$. It also includes an overlap-conditioned interface that
can preserve a committed action prefix and generate only a successor suffix.

The current LIBERO results do not support a claim that the speedup is obtained
without a loss of task performance. Under a two-A800 training setup, three
drifting seeds show $64.0\pm4.0\%$ success on Spatial, $52.0\pm1.0\%$ on Goal,
and $26.0\pm2.6\%$ on Long. The seed stability indicates that the degradation
is systematic, while the cause remains unresolved.

The most useful next steps are controlled rather than promotional. They
include matching the training and evaluation protocol for all baselines,
removing the minibatch dependence of the geometry estimate, testing whether
the query sample should be excluded from the neighborhood, reducing the number
of actions executed per chunk, and evaluating the overlap path in an
asynchronous rollout. A one-step policy should be judged not only by its
action-head latency but by its speed--success frontier and by whether the
latency reduction enables more frequent closed-loop feedback.

\appendices
\section{Checkpoint and Evaluation Inventory}
\label{app:inventory}
\sloppy

The primary multi-seed drifting checkpoints are:
\begin{itemize}
    \item \path{Xihe666/drif_ov_libero_20k_seed1_0821}
    \item \path{Xihe666/drif_ov_libero_20k_seed42_0821}
    \item \path{Xihe666/drif_ov_libero_20k_seed1000_0822}
\end{itemize}
These runs were trained on two NVIDIA A800 GPUs. The documented seed-1000
configuration uses 20k steps, batch size 32 per GPU, and bfloat16 precision.
The seed-1 and seed-42 runs are evaluated as part of the same multi-seed study.

The matched GR00T seed-comparison checkpoints are:
\begin{itemize}
    \item \path{Xihe666/gr00t_n17_libero_20k_seed1}
    \item \path{Xihe666/gr00t_n17_libero_20k_seed42}
\end{itemize}

The legacy single-run checkpoint is:
\begin{itemize}
    \item \path{Xihe666/drif_ov_libero0809}
\end{itemize}
It was trained with four A800 GPUs, batch size 256 per GPU, and 15k steps.
It is retained only to document the development history and is excluded from
the primary comparison.

The raw latency CSV files and summary JSON files are stored under
\texttt{Experiment-result/latency\_bench} and
\texttt{Experiment-result/LIBERO\_latency\_stats}. The aggregate summary used
to construct the tables in this report is stored under
\texttt{Experiment-result/LIBERO\_summary}.

\section{Excluded Measurements}
\label{app:excluded}

The archived pi0.5 evaluation reports $0.0\%$ success for Spatial, Object,
Goal, and Long. The timing CSV and summary JSON also disagree for several
tasks. Because a uniform zero success rate is more likely to reflect an
evaluation or action-mapping failure than a meaningful policy result, the
pi0.5 table is excluded from the primary comparison.

The archived object summary has no drifting success value. We therefore
report the object latency measurement but leave object success as
\texttt{N/A}. No missing success value is replaced by a legacy run or by the
mean of other suites.

\bibliographystyle{IEEEtran}
\bibliography{refs}

\end{document}